\documentclass[letterpaper]{article} 
\usepackage[draft]{aaai2026}  
\usepackage{times}  
\usepackage{helvet}  
\usepackage{courier}  
\usepackage[hyphens]{url}  
\usepackage{graphicx} 
\usepackage{natbib}  
\usepackage{caption} 
\usepackage{algorithm}
\usepackage{times}
\usepackage{epsfig}
\usepackage{graphicx}
\usepackage{amsmath}
\usepackage{amssymb}
\usepackage{algpseudocode}
\usepackage{booktabs}
\usepackage{siunitx} 
\usepackage{pifont}
\usepackage{makecell}
\usepackage{xcolor}

\usepackage{newfloat}
\usepackage{listings}
\DeclareCaptionStyle{ruled}{labelfont=normalfont,labelsep=colon,strut=off} 
\floatstyle{ruled}
\newfloat{listing}{tb}{lst}{}
\floatname{listing}{Listing}
\title{\cache: Adaptive KV Caching in Multi-Scale Visual Autoregressive Transformers}
\author{
    Boxun Xu, Yu Wang, Zihu Wang, Peng Li\\
}
\affiliations{
    University of California, Santa Barbara\\
    \{boxunxu, yu95, zihu\_wang, lip\}@ucsb.edu
}

\usepackage{bibentry}

\begin{document}
\newcommand{\cache}{\texttt{AMS-KV}}
\newcommand{\ourmethod}{\texttt{AMS-KV}\space}
\newcommand{\ourmethodnospace}{\texttt{AMS-KV}}
\newcommand{\nospace}{\unskip}
\newcommand{\blfootnote}[1]{%
  \begingroup
  \renewcommand\thefootnote{}%
  \footnote{#1}%
  \addtocounter{footnote}{-1}%
  \endgroup
}

\maketitle
\blfootnote{This paper has been accepted for publication in the Proceedings of the AAAI Conference on Artificial Intelligence (AAAI-26).}
\begin{abstract}
Visual autoregressive modeling (VAR) via next-scale prediction has emerged as a scalable image generation paradigm. While Key and Value (KV) caching in large language models (LLMs) has been extensively studied, next-scale prediction presents unique challenges, and KV caching design for next-scale based VAR transformers remains largely unexplored. A major bottleneck is the excessive KV memory growth with the increasing number of scales—severely limiting scalability. 
Our systematic investigation reveals that: (1) Attending to tokens from local scales significantly contributes to generation quality (2) Allocating a small amount of memory for the coarsest scales, termed as condensed scales, stabilizes multi-scale image generation (3) Strong KV similarity across finer scales is predominantly observed in cache-efficient layers, whereas cache-demanding layers exhibit weaker inter-scale similarity.
Based on the observations, we introduce \cache, a scale-adaptive KV caching policy for next-scale prediction in VAR models. \ourmethod prioritizes storing KVs from condensed and local scales, preserving the most relevant tokens to maintain generation quality. 
It further optimizes KV cache utilization and computational efficiency identifying cache-demanding layers through inter-scale similarity analysis.
Compared to the vanilla next-scale prediction-based VAR models, \ourmethod reduces KV cache usage by up to 84.83\% and self-attention latency by 60.48\%. 
Moreover, when the baseline VAR-d30 model encounters out-of-memory failures at a batch size of 128, \ourmethod enables stable scaling to a batch size of 256 with improved throughput. 

\end{abstract}


\section{Introduction}
\begin{figure*}
    \centering
    \includegraphics[width=0.99\linewidth,trim=40mm 70mm 50mm 20mm,clip]{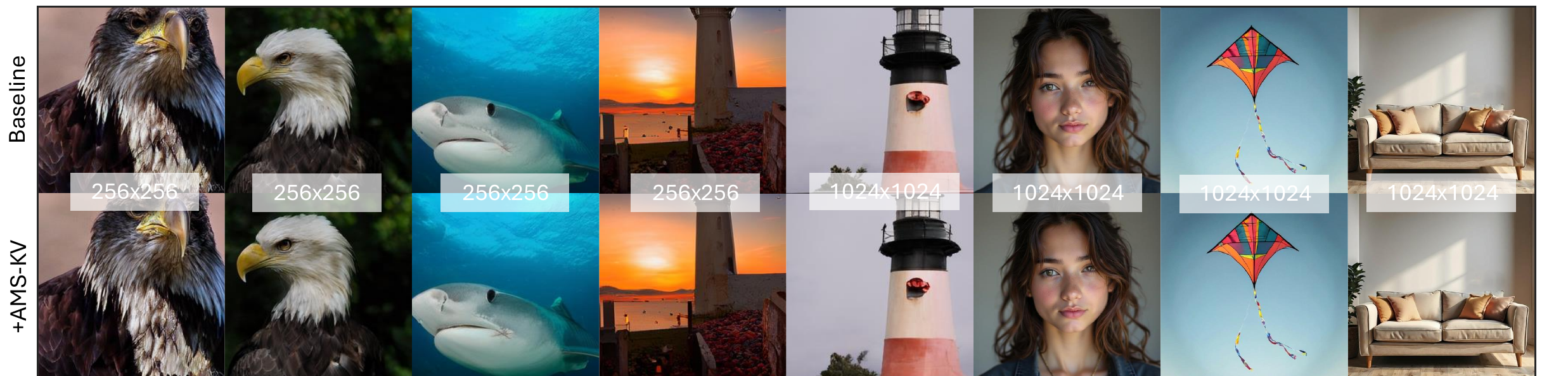}
    \caption{\textbf{Top:} The $256\times 256$ images generated by VAR-d30\cite{var} and $1024\times 1024$ images generated Infinity-2B\cite{infinity}. \textbf{Bottom:} generated using tuning-free \ourmethod  with 4.7$\times$ less KV Cache Memory consumption.}
    \label{fig:examples}
\end{figure*}

Autoregressive (AR) models have demonstrated remarkable success in natural language processing (NLP), underpinning the development of large language models (LLMs). Scaling laws \cite{gpt4, scalinglaw} reveal that AR models benefit significantly from increased model size, larger datasets, and expanded computational resources, making them the foundation of the state-of-the-art NLP systems. However, their efficiency remains a critical challenge, particularly due to the large memory overhead introduced by storing key-value (KV) caches during inference.

Encouraged by the success of AR models in NLP, efforts have been made to extend AR modeling to vision tasks, leading to promising results in image and video generation.  Among these, VQ-VAE \cite{vqvae} introduced quantized latent tokens to greatly improve computational efficiency over pixel-level AR modeling \cite{pixelcnn, conditionalpixelcnn, generative_pretraining_pixel}. Follow-up works have enhanced quantization quality via hierarchical structures \cite{vqvae2}, residual quantization \cite{rqvae}, or generative adversarial frameworks \cite{vqgan}. 
Nevertheless, token-level AR models have historically underperformed compared to diffusion models \cite{diffusionoriginal, diffusionsongyang, dit, stablediff3}, which however require multiple expensive iterative sampling steps.

The recently developed visual autoregressive (VAR) models \cite{var} have adopted GPT-style next-token prediction through next-scale prediction, enabling a new coarse-to-fine progressive image generation paradigm and surpassing diffusion-based generative models in performance. VAR models offer a more structured and controllable generative process \cite{var, litevar, collabvar}, and have been successfully generalized to tasks like text-to-image synthesis \cite{switti, varclip, star, infinity}, 3D reconstruction \cite{vat, sar3d}, and multi-modal generation \cite{tokenflow}. However, practical scalability of VAR remains a challenge.

Model efficiency has been extensively studied in NLP, with techniques such as pruning \cite{llmpruner, compactlanguagemodelspruning, llmcompressionsurvey1}, quantization \cite{deepcompression, awq, displlm, compressionsurvey1}, and cache optimization \cite{streamingllm, h2o, lckv, yoco, kivi, kvquant, reducingtransformerkeyvaluecache, minicache, skiplayerattn}.
In vision, techniques of improving diffusion model efficiency have also been well developed, including reducing the number of sampling steps \cite{lcm, progressivedistillationfastsampling} and accelerating solvers \cite{dpmsolver, dpmsolverpp}.
However, similar optimizations for VAR remain relatively scarce \cite{litevar, collabvar}. 

Notably, the progressive nature of VAR significantly extends the overall sequence length compared to conventional AR models. This results in excessive memory usage, primarily due to the KV cache, a fundamental bottleneck limiting the scalability of VAR models. Unlike LLMs, which only maintain a single sequence of KV cache storage, VAR models operate across multiple scales, requiring progressively larger cache storage as the generation process unfolds. This multi-scale dependency significantly amplifies the KV cache footprint, making VAR models memory-intensive and challenging to scale to higher resolutions or more complex scenes.

However, addressing KV cache efficiency in VAR models has received little attention, despite its critical role in enabling scalable deployment. 
In this work, we bridge this gap by systematically investigating KV cache redundancy in VAR models and proposing an Adaptive Multi-Scale KV caching strategy, called \cache, to significantly reduce memory consumption while maintaining model performance. Our main contributions are:

\begin{itemize}
    \item We conduct in-depth studies of KV caching behavior across different \textbf{scales} and \textbf{layers} in VAR models, revealing key insights into non-uniform cache requirements, scale-wise importance and layer-wise cache preference.
    \item We propose a novel KV caching policy to dynamically allocate cache storage based on scale types across layers with heterogeneous cache demands. This enables substantial memory savings as shown in Figure~\ref{fig:examples} and supports larger batches while keeping generation quality.
    \item We perform extensive experiments and ablation studies to validate the performance of \cache, demonstrating significantly improved memory efficiencies.
\end{itemize}

By addressing KV cache inefficiencies in VAR models, \ourmethod facilitates the scalable deployment of AR-based vision models, enabling broader applicability in real-world scenarios.

\section{Related Works}

\subsection{Autoregressive Vision Modeling}

Autoregressive (AR) models have achieved remarkable success in natural language processing (NLP), leading to the development of influential foundation models \cite{gpt2,llama,jiang2023mistral7b,deepseekai2025deepseekv3technicalreport}. Recently, significant effort has been dedicated to adapting AR modeling to vision tasks.

Early attempts focused on pixel-level AR modeling \cite{pixelcnn,conditionalpixelcnn,generative_pretraining_pixel}, demonstrating feasibility but facing scalability challenges in generating high-resolution images. To overcome this, VQ-VAE \cite{vqvae} introduced quantized latent tokens, greatly improving computational efficiency. Follow-up studies have enhanced quantization quality via hierarchical structures \cite{vqvae2}, residual quantization \cite{rqvae}, or generative adversarial frameworks \cite{vqgan}. Parallel developments directly leverage transformer architectures for visual modeling \cite{vit,vitvqgan}, opening opportunities for downstream tasks such as text-to-image generation \cite{parti,dalle,aim,mars}.

Despite these advancements, token-level AR models have typically underperformed diffusion models \cite{diffusionoriginal, diffusionsongyang, dit, uvit, stablediff3}. Until recently, VAR \cite{var} introduced a novel AR paradigm based on next-scale prediction, achieving superior performance compared to diffusion models. This approach has successfully generalized to tasks like text-to-image synthesis \cite{switti, varclip, star, infinity}, 3D reconstruction \cite{vat, sar3d}, and multi-modal generation \cite{tokenflow}.

\begin{figure*}
    \centering
    \includegraphics[width=0.99\linewidth,trim=65mm 45mm 5mm 26mm,clip]{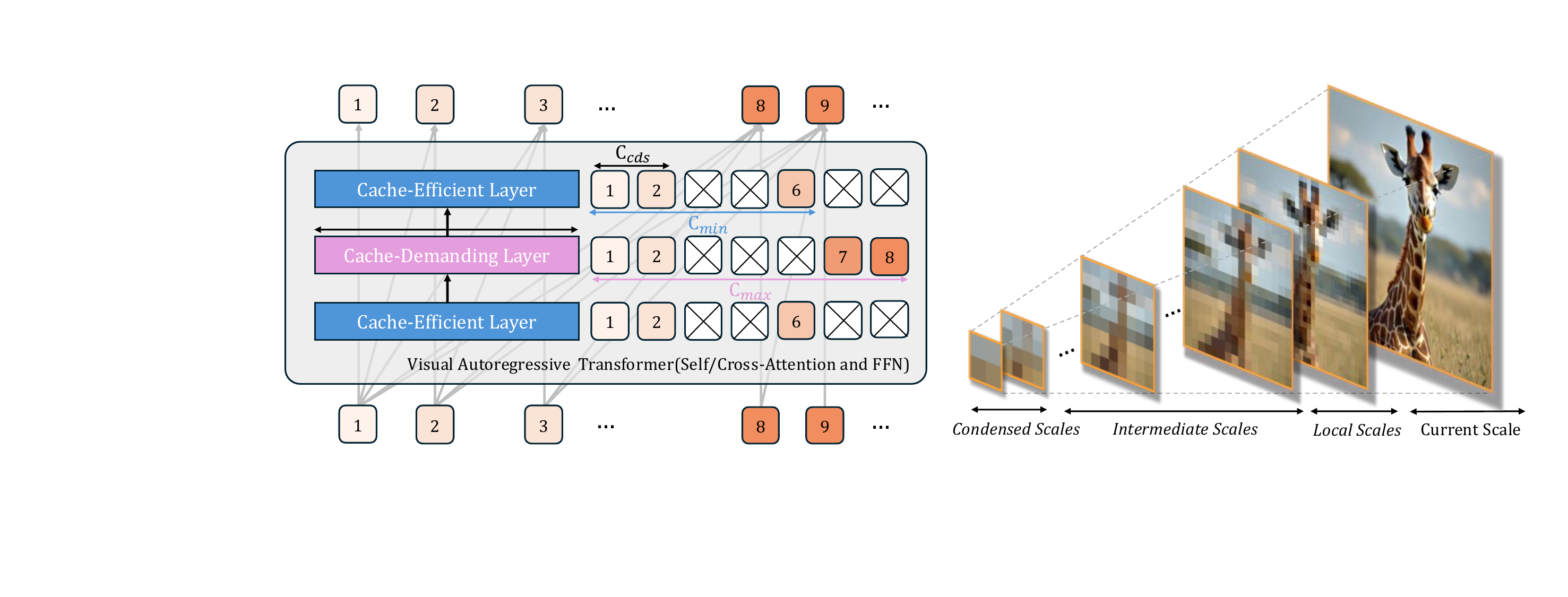}
    \caption{Overview of Adaptive Multi-Scale KV Caching (\ourmethodnospace) for Visual Autoregressive Modeling.}
    \label{fig:amskv_overview}
\end{figure*}

\subsection{Efficient Language and Vision Models}
The success of autoregressive (AR) models in NLP has driven extensive research on scaling laws \cite{gpt4, scalinglaw}, revealing that performance scales effectively with increased model size, data, and computation. However, the computational cost remains a major challenge for practical deployment.

To address redundancy in over-parameterized networks for language modeling, compression techniques such as pruning \cite{llmpruner, compactlanguagemodelspruning, llmcompressionsurvey1}, quantization \cite{deepcompression, awq, displlm, compressionsurvey1}, and knowledge distillation\cite{yin2024one,yin2024improved} have been explored. For AR models, attention mechanisms introduce additional bottlenecks due to large key-value (KV) caches. Recent works mitigate this by discarding less critical cache entries \cite{streamingllm, h2o, lckv, yoco}, applying quantization \cite{kivi, kvquant}, or enabling cache sharing across layers \cite{reducingtransformerkeyvaluecache, minicache, skiplayerattn}.

In vision models, research primarily focuses on diffusion-based architectures by optimizing sampling efficiency via reduced steps \cite{lcm, progressivedistillationfastsampling} and faster solvers \cite{dpmsolver, dpmsolverpp}. In contrast, efficiency improvements for VAR models remain largely underexplored, with only a few initial efforts \cite{litevar, collabvar}. 


\section{Preliminaries}

\subsection{Visual Autoregressive Modeling (VAR)}
\begin{figure}
    \centering
    \includegraphics[width=\linewidth, trim=80mm 50mm 170mm 35mm, clip]{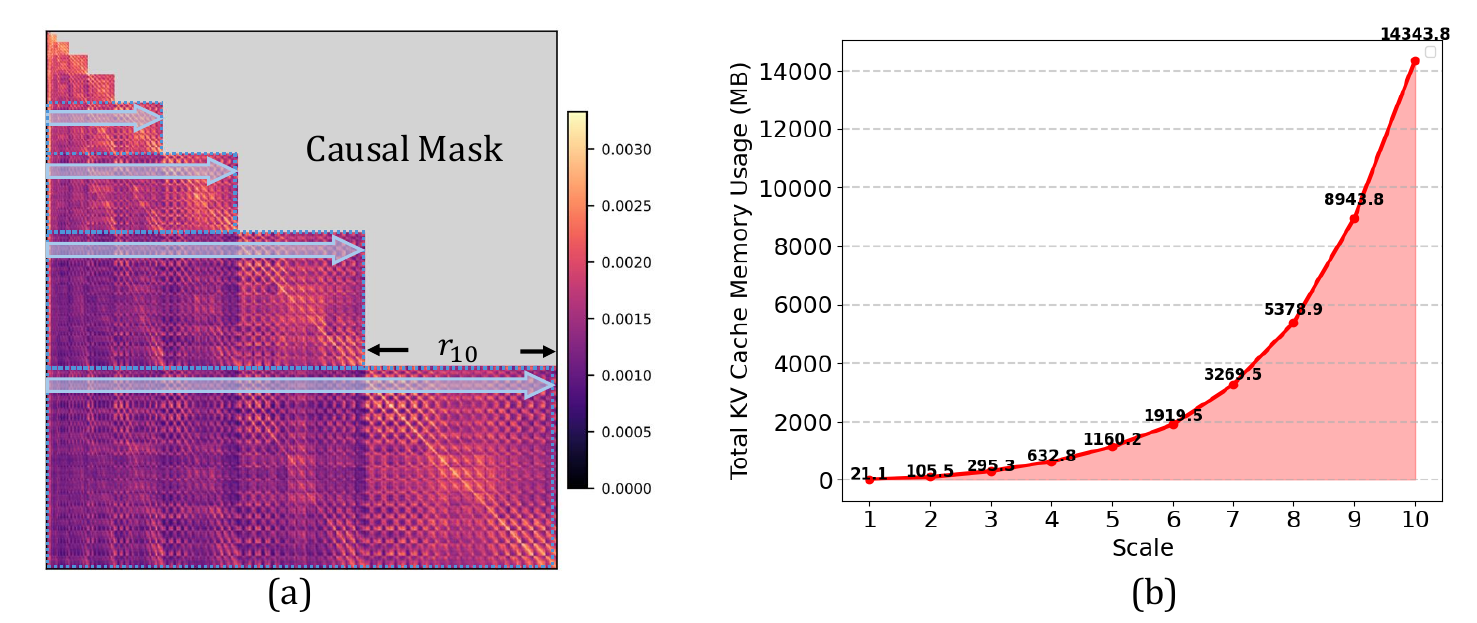}
    \caption{(a) The growth of attention map. (b) Memory usage of VAR across scales during unconditional image generation on an NVIDIA A100-80G (Batch Size=50)}
    \label{fig:attn_overview}
\end{figure}

Visual Autoregressive (VAR) models build upon the principles introduced by VQ-VAE \cite{vqvae}, but further extends the standard next-token prediction approach to next-scale prediction. Specifically, an input image is first encoded into a multi-scale hierarchy of discrete tokens using a Vector-Quantized Variational Autoencoder (VQ-VAE):
\begin{equation}
    f = \mathcal{E}(im), \quad R := (r_1, r_2, \dots, r_K) = \mathcal{Q}(f),
\end{equation}
where \( f \in \mathbb{R}^{h\times w\times d} \) denotes the encoded feature map. Each discrete scale \( r^k \in [V]^{h^k\times w^k} \)\footnote{For simplicity, we let $h^k=w^k$ in later sections, e.g., we abbreviate the spatial size of a scale of $4\times 4$ to 4.} is obtained through a residual quantizer \(\mathcal{Q}\) \cite{rqvae} that utilizes a shared codebook.

A decoder-only transformer architecture factorizes the joint likelihood of these scales autoregressively:
\begin{equation}
    p(r_1, r_2, \dots, r_K) = \prod_{k=1}^{K} p(r_{k}\mid r_{<k}),
\end{equation}
which sequentially predicts coarse scales with smaller spatial dimensions to larger fine-grained scales.  This next scale prediction allows VAR to progressively capture information from general structures to finer details, and achieves outstanding performance in visual representation learning and image synthesis tasks.

\subsection{Memory Bottleneck in Efficient VAR}
Despite its strong performance, VAR models encounter significant memory bottlenecks during next-scale prediction. Standard VAR implementations typically use scales with spatial dimensions that grow non-linearly, such as \([1\times 1], [2\times 2], \dots, [16\times 16]\). This quadratic increase in spatial dimension results in rapidly growth of computational complexity for the attention mechanism, as illustrated in Figure~\ref{fig:attn_overview} (a). To mitigate this complexity, VAR adopts the Key-Value (KV) cache technique, widely used in large language models (LLMs), which efficiently caches previously computed key and value matrices to avoid redundant computations \cite{llama}. Nevertheless, due to the quadratic growth of each subsequent scale's spatial size, the size of newly computed keys and values also expands quadratically, ultimately resulting in a cubic growth of the overall KV cache size. This rapid growth can quickly surpass memory limitations, particularly during predictions at later scales, as illustrated in Figure~\ref{fig:attn_overview}(b).






\section{Methodology}

The expanding KV cache in VAR models presents a significant computational and memory challenge. This necessitates an efficient KV cache management strategy that can prune redundant information without compromising generation quality.
A critical step toward such a strategy is to understand the roles of different scales existing in different layers. 
We have observed that due to the sequential nature of scale generation in VAR, a query at a certain stage often exhibits similar attention patterns across multiple distinct scales as shown in Figure~\ref{fig:attn_overview}. This suggests a high degree of semantic similarity, particularly in later stages, and reveals substantial redundancy in the information stored across scales. 

To address this, we identify two key properties:  
(1) Not all scales need to be stored and they do not contribute equally to generation, a property we refer to as \textbf{scale-wise importance}.  
(2) Storing even the most important scales is still challenging due to the quadratic size complexity. However, we observe that KV cache requirements are heterogeneous across different layers. We term this as \textbf{layer-dependent cache preference} and classify layers as either cache-efficient or cache-demanding.



\subsection{Not All Scales Are Equally Important}\label{subsec:scale}

Inspired by the notion of token importance in LLMs, we hypothesize that different scales contribute unequally to the generation process. To analyze this non-uniformity, we examine the cross-scale attention—the extent to which the final scale (e.g., $r_{10}$ in VAR models) attends to tokens from preceding scales across different heads. We observe that scales influence the generation process differently, contributing unevenly to next-scale predictions.

As we concern scale-wise patterns and token counts vary across different scales, directly comparing raw accumulated cross-scale attention values for each previous scale is not meaningful. 
To account for this, we introduce \textit{attention density}, defined as $d_{i\leftarrow j}
= \frac{1}{T_i} \sum_{p=1}^{T_j} \sum_{q=1}^{T_i} A^{(j\to i)}_{p,q}$,
where $A^{(j\to i)} \in \mathbb{R}^{T_j \times T_i}$ denotes the cross-scale attention matrix from scale $j$ to $i$, and $T_i$ is the token number of scale $i$.


\begin{figure}[h]
    \centering
    \includegraphics[width=0.98\linewidth, trim=20mm 45mm 75mm 20mm, clip]{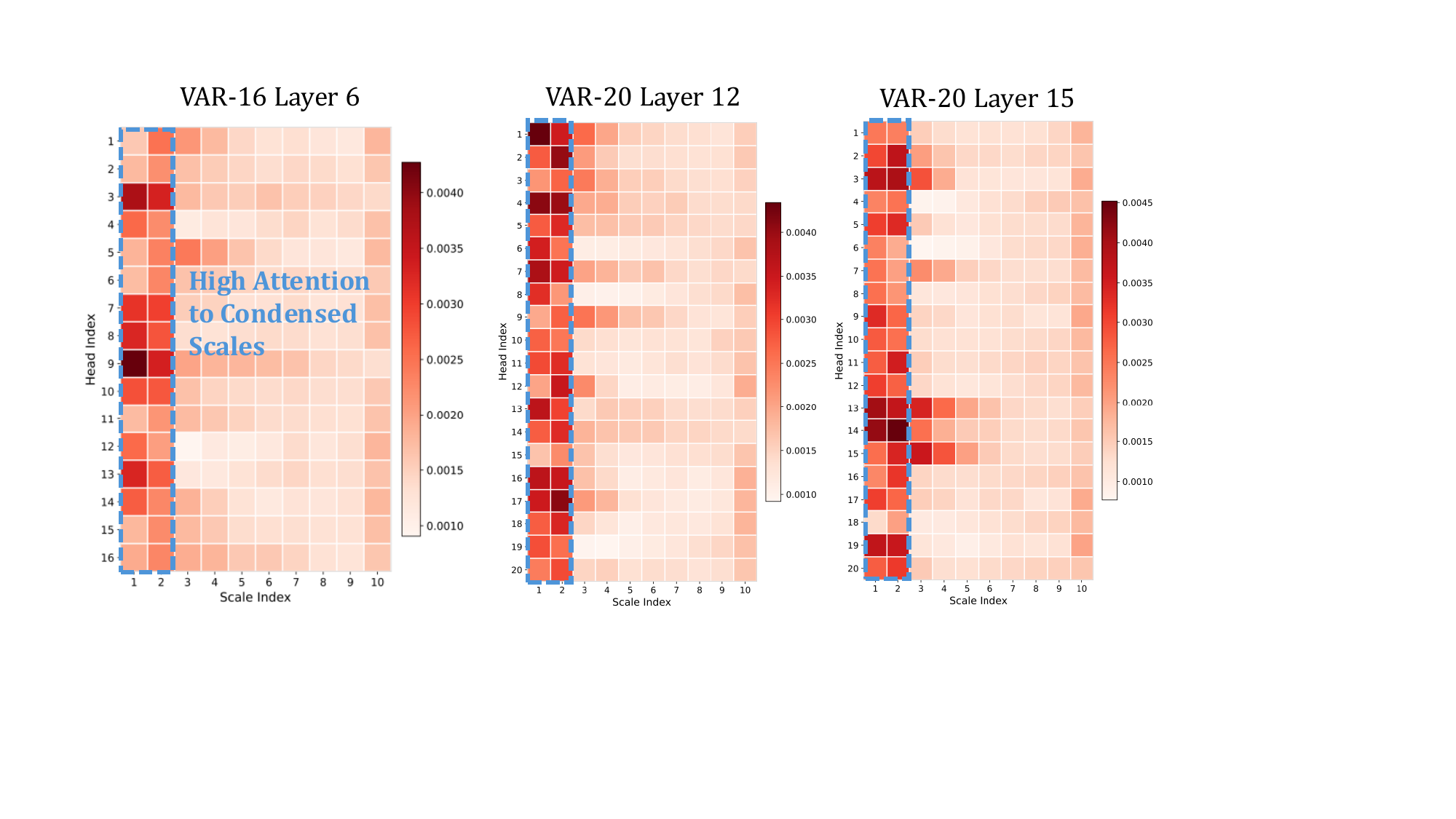}
    \caption{Visualization of the attention density across heads and scales when generating the last scale $r_{10}$.}
    \label{fig:attn_density}
\end{figure}

Interestingly, as illustrated in Figure~\ref{fig:attn_density}, VAR models exhibit a characteristic pattern: the initial scales, despite having fewer tokens, exhibit the highest attention density. This implies that storing such scales are most high cost performance. The attention density decreases across intermediate scales before rising again at the local scales. This trend suggests that both early and late scales play a critical role in generation, while intermediate scales contribute less significantly.
Based on this observation, we categorize scales into three groups:  
Condensed scales, referring to the first two scales; Local scales, comprising the most recent scales preceding the current scale; and Intermediate scales which include all remaining scales not covered by the above two groups.

\begin{table}[h]
    \centering
    \caption{Comparison of Generation Quality When Removing KV Caches from Different Scales.}
    \setlength{\tabcolsep}{0.8mm}{
    \begin{tabular}{l|cc|cc}
        \toprule
        \textbf{Config.} & \textbf{FID$\downarrow$} & \textbf{IS$\uparrow$} & \textbf{Precision$\uparrow$} & \textbf{Recall$\uparrow$} \\
        \midrule
        Baseline(No removal) & 2.07 & 336.15 & 0.82 & 0.58 \\
        Remove Intermediate & 2.12 & 329.60 & 0.82 & 0.59 \\
        Remove Local  & 3.04 & 295.14 & 0.76 & 0.60 \\
        Remove Condensed  & 2.48 & 298.35 & 0.80 & 0.59 \\
        \bottomrule
    \end{tabular}
    }
    \label{tab:pls_effect}
\end{table}

\begin{figure}[h]
    \centering
    \includegraphics[width=0.99\linewidth]{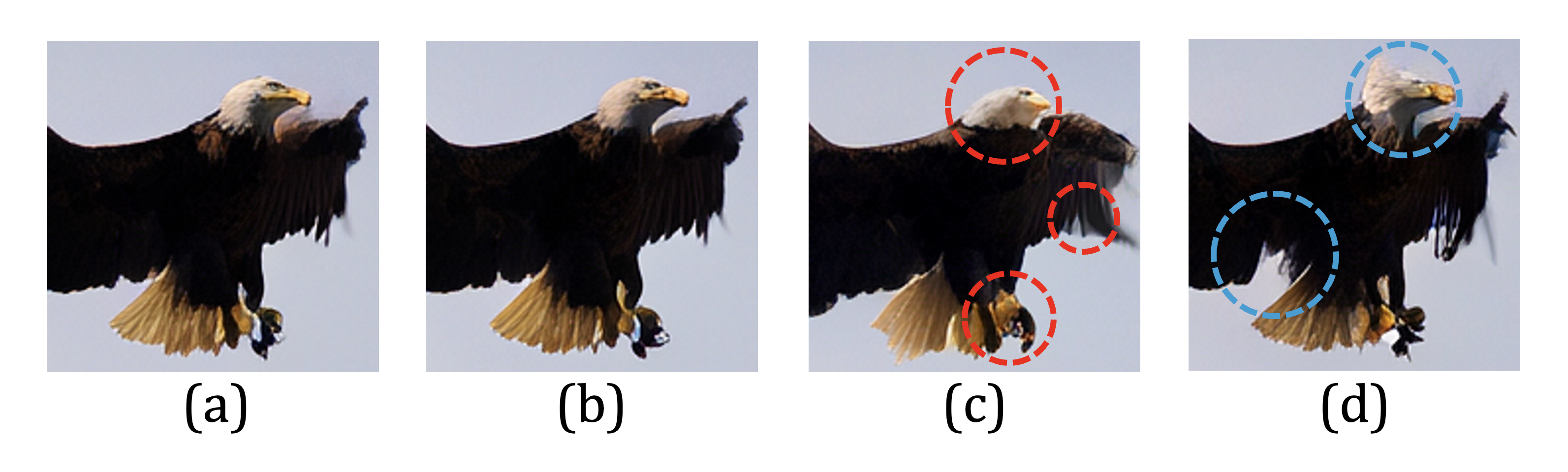}
    \caption{An eagle generated by VAR-d30 with (a) full KV cache; (b) intermediate scales removed; (c) local scales removed; (d) condensed scales removed. Red circles highlight missing fine details; blue circles indicate distorted regions.}
    \label{fig:scale-analysis}
\end{figure}

To further assess the impact of these scale groups, we conduct an experiment that selectively removes KV caches for different scale groups during image generation. The results reveal distinct effects, as shown in Figure~\ref{fig:scale-analysis}. Firstly, condensed scales serve as an anchor for modeling the global structural foundation, removing these scales disrupts the overall composition of the generated images. Secondly, local scales refine the details. Without them, synthesized images become blurry and lack sharpness. Finally, intermediate scales have minimal impact when excluded, indicating redundancy in the KV cache for these scales. 
We also compare the image generation quality between the above three settings in Table~\ref{tab:pls_effect}, which further conforms our conclusion.

These findings highlight that memory allocation for KV caches in VAR models can be optimized by prioritizing the condensed and local scales while reducing storage for intermediate scales, thereby reducing overall memory overhead without degrading generation quality.


\subsection{Not All Layers are Cache-Demanding}
\label{sec:4-2}

In decoder-only transformer architectures with KV cache support~\cite{llama, infinity}, the cache budget is typically allocated uniformly across all decoder layers—each layer is assigned the same maximum cache size. 
However, this default strategy is suboptimal for VAR models. Our analysis reveals that cache-demanding layers—those with low inter-scale KV similarity—play a critical role in feature expansion and structural propagation, and thus require larger KV caches to preserve essential generation context. In contrast, cache-efficient layers exhibit high KV similarity across scales and mainly perform refinement over already-formed representations, making them more resilient to reduced cache allocation.

Moreover, uniformly storing local scales across all layers becomes impractical, as the token count within each scale grows quadratically. 
Under a fixed total cache budget, we observe that not all layers should be treated equally in cache allocation.
Specifically, prioritizing cache resources for cache-demanding layers, to store more local scales, helps preserve generation quality.
In contrast, assigning minimal cache to cache-efficient layers, limiting them to a few local or even sub-local scales can further reduce memory usage without harming performance. Here, “sub-local” scales refer to earlier scales that are even smaller than the local scales.
This motivates a non-uniform, similarity-aware cache allocation strategy that aligns cache resources with the functional roles of different decoding layers in VAR models.

\begin{figure}
    \centering
    \includegraphics[width=0.98\linewidth, trim=20mm 2mm 15mm 2mm, clip]{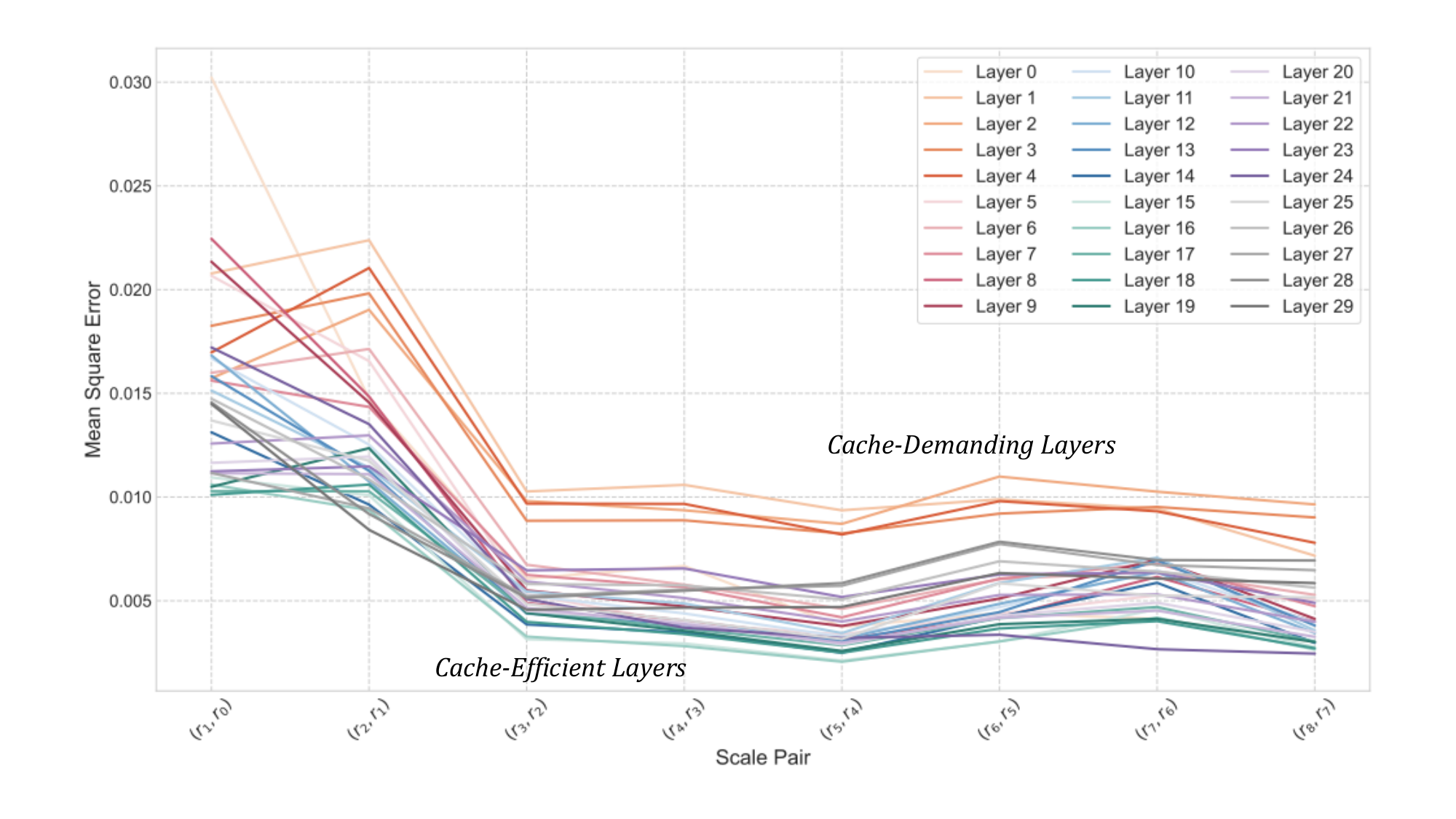}
    \caption{Discrepancy between adjacent-scale keys across layers in the VAR-d30 model.}
    \label{fig:Distance}
\end{figure}

To validate the observation, we conduct an illustrative experiment under a constrained total cache budget. This total budget allows one-sixth of layers assigned with a large layer cache capacity of $C_{\max}$, sufficient to store the two most recent scales and condensed scales while assigning a smaller layer capacity of $C_{\min}$ to the remaining five-sixths layers, which can exactly store the second most recent and condensed scales. These caches are distributed across the decoding layers of a VAR-d30 model.
We compare the following two cache allocation strategies under the same total cache budget with the same number of layers using larger caches as described above. 
In the uniform allocation, \texttt{S1}, the layers with larger caches are uniformly distributed across model depth. In contrast, in the similarity-aware allocation, \texttt{S2}, the larger caches are assigned to the layers exhibiting the lowest inter-scale similarity, as defined in Equation~\ref{eq:scale_similarity} and identified in Figure~\ref{fig:Distance}.

We evaluate the impact of these two strategies on image generation quality as  summarized in Table~\ref{tab:layer_effects}. The results further confirm that low-similarity layers, as cache-demanding layers, are crucial for feature construction and require larger cache capacity. Reducing the cache size in these layers leads to a significant degradation in performance, whereas cache-efficient layers exhibit greater robustness to cache compression.


\begin{table}[h]
    \centering
    \caption{Comparison of generation quality under two allocation strategies, \texttt{S1} (uniform) and \texttt{S2} (similarity-aware).}
    \setlength{\tabcolsep}{2.6mm}{
    \begin{tabular}{l|cc|cc}
        \toprule
        \textbf{Policy} & \textbf{FID$\downarrow$} & \textbf{IS$\uparrow$} & \textbf{Prec.$\uparrow$} & \textbf{Rec.$\uparrow$} \\
        \midrule
        \texttt{S1} (Uniform)       & 6.65 & 223.65 & 0.71 & 0.60 \\
        \texttt{S2} (Sim-aware)         & 3.37 & 267.71 & 0.76 & 0.61 \\
        \bottomrule
    \end{tabular}
    }
    \label{tab:layer_effects}
\end{table}


\subsection{\ourmethodnospace: Adaptive Multi-Scale KV Caching}
The proposed KV caching policy, \ourmethodnospace, is motivated by the aforementioned observations of scale-wise importance and layer-dependent cache preference.
First, \ourmethod retains the condensed scales while rolling local scales. Second, it employs a similarity-driven approach to classify layers into \textbf{cache-demanding} and \textbf{cache-efficient} layers, enabling cache-demanding layers to store larger and more local scales.

As illustrated in Figure~\ref{fig:amskv_overview}, each layer's KV cache in \ourmethod operates independently and is governed by three key hyperparameters: $C_{\min}$, $C_{\max}$ and $C_{\text{cds}}$. Specifically, $C_{\max}$ represents the expanded capacity allocated to cache-demanding layers, which is the maximum cache size for a layer; $C_{\min}$ defines the reserved capacity of cache-efficient layers which is the minimum cache size for a layer; $C_{\text{cds}}$ specifies the number of condensed scales retained within the KV cache during rolling.

The update rule is as illustrated in Algorithm~\ref{alg:kv_cache}. 
Each layer is initialized as cache-efficient layer with its cache budget $C_{bgt}$ set to $C_{\min}$.
During finer-grained rolling of KV pairs, the KV cache always prioritizes storing the condensed scales, which are locked due to their high cross-scale attention, as illustrated in Figure~\ref{fig:attn_density}.
In addition, the KV cache dynamically appends local finer scales by evicting one or multiple older intermediate scales. 
To manage this process, we introduce the \underline{C}ondensed \underline{L}east \underline{R}ecently \underline{U}sed ($\texttt{CLRU}$) policy, a FIFO-based eviction strategy that removes older scales while ensuring that condensed scales are retained, thereby maintaining the cache size within the budget $C_{bgt}$, as shown in Algorithm~\ref{alg:clru}.

If the initial cache budget $C_{bgt} = C_{\min}$ is exhausted, \ourmethod invokes a similarity-guided mechanism to determine whether the current layer should be classified as a \textbf{cache-demanding} layer.  
To quantify inter-scale similarity, the keys from the $(i{-}1)$-th scale, denoted as $k_{i-1}$, are resized via 2D interpolation to match the spatial resolution of $k_i$, resulting in $\hat{k}_{i-1}$. The similarity score is then computed as the negative $\ell_2$ distance:
\begin{equation}
\label{eq:scale_similarity}
\text{S}(k_i, \hat{k}_{i-1}) = -\lVert k_i - \hat{k}_{i-1} \rVert_2.
\end{equation}
If this similarity falls below a predefined threshold $\theta$, the layer is classified as cache-demanding.

The cache extends its cache budget $C_{bgt}$ only when two conditions are satisfied. 
First, the current cache size budget $C_{bgt}$ can be further increased to $C_{\max}$. Second, the layer is identified as a cache-demanding layer.
If the both conditions hold, the layer's cache size budget is expanded from $C_{\min}$ to $C_{\max}$ by allocating the additional memory $(C_{\max}-C_{\min})$. Otherwise, the cache budget remains fixed at $C_{\min}$. Note that $C_{\min}>>C_{\text{cds}}$.

In this manner, the cache-efficient layers in \ourmethod are processed with $C_{\min}$, while the detected cache-demanding layers are assigned a cache capacity of $C_{\max}$, particularly as the cache size of scale $r_{i}$ increases.

\begin{algorithm}
\caption{Adaptive Multi-Scale KV Caching}
\label{alg:kv_cache}
\begin{algorithmic}[1]
\Procedure{\textsc{Cache}}{$\{(k_i, v_i)\}_{i=1}^{N_{\text{scales}}}, C_{\min},  C_{\max}, C_{\text{cds}}$}
\State{$\mathcal{C}_{0} \gets \emptyset$} 
\State{$C_{bgt}\gets C_{\min}$}  \Comment{Initialized as Cache-Efficient layers}
\For{$i = 1 \rightarrow N_{\text{scales}}$}
    \If{$ |(k_i, v_i)| > C_{\max}$}
        \State{Continue}
    \EndIf
    \State $\mathcal{C}_{i} \gets \mathcal{C}_{(i-1)} \cup \{(k_i, v_i)\}$ \Comment{Append $i$-th scale}
    \If{$|\mathcal{C}_{i}| > C_{bgt}$}
        \If{$C_{bgt} == C_{\min}$ \& $\text{S}(k_i, \hat{k}_{(i-1)}) < \theta$}
            \State $C_{bgt} \gets C_{\max}$ \Comment{Expand Cache Budget}
        \EndIf
        \If{$|(k_i, v_i)| > C_{bgt}$}
            \State{$\mathcal{C}_{i} \gets \mathcal{C}_{(i-1)}$ and Continue}
        \EndIf
        \State $\mathcal{C}_{i} \gets \texttt{CLRU}(\mathcal{C}_{i}, C_{bgt}, C_{\text{cds}})$ \Comment{Rolling}
    \EndIf
\EndFor
\EndProcedure
\end{algorithmic}
\end{algorithm}

\section{Experiments}

\subsection{End-to-End Results}\label{sec:main_exp}
\begin{table*}[ht]
    \centering
    \caption{Performance Comparison between different model configurations in VAR model families. $^{1}$ indicates a configuration of CFG=1.5 and top-k=900; $^{2}$ indicates a batch size of 25 on specific models. $\lozenge$ denotes an approximately KV Cache requirement of 117GB, which causes Out-of-Memory(\texttt{OOM}).}
    \setlength{\tabcolsep}{2.2mm}{
    \begin{tabular}{l|c|cc|cc|cc|c}
        \toprule
        \textbf{Model Config.} & \textbf{Image Res.} & \textbf{\#Para.} & \textbf{\#Dim} & \textbf{FID$\downarrow$} & \textbf{IS$\uparrow$} & \textbf{Prec.$\uparrow$} & \textbf{Rec.$\uparrow$} & \textbf{KV Cache Size} \\
        \midrule
        VAR-d30\cite{var} & 256x256 & 2.0B & 1920 & 2.07 & 336.15 & 0.82 & 0.58 & 22.41GB \\
        VAR-d30$^{1}$\cite{var} & 256x256 & 2.0B & 1920 & 2.03 & 305.19 & 0.82 & 0.59 & 22.41GB \\
        VAR-d30 (+\cache) & 256x256 & 2.0B & 1920 & 2.09 & 330.50 & 0.82 & 0.58 & 4.77GB (78.72\%$\downarrow$) \\
        VAR-d30$^{1}$(+\cache) & 256x256 & 2.0B & 1920 & 2.09 & 301.19& 0.82 & 0.58 & 4.77GB (78.72\%$\downarrow$) \\
        \midrule
        VAR-d36$^{2}$\cite{var} & 512x512 & 2.4B & 2304 & 3.26 & 331.13 & 0.82 & 0.34 & 53.15GB \\
        VAR-d36\cite{var} & 512x512 & 2.4B & 2304 & - & - & - & - & \texttt{OOM}$^{\lozenge}$ \\
        VAR-d36$^{2}$ (+\cache) & 512x512 & 2.4B & 2304 & 3.31 & 331.18 & 0.82 & 0.34 & 15.26GB (71.29\%$\downarrow$) \\
        VAR-d36 (+\cache) & 512x512 & 2.4B & 2304 & - & - & - & - & 30.52GB \\
        \bottomrule
    \end{tabular}
    }
    \label{tab:model_performance}
\end{table*}

\paragraph{Experimental Setup}
We conduct experiments using pretrained Visual Autoregressive (VAR) models ($d=30, 36$) obtained from \cite{var} and Infinity-2B\cite{infinity}. The primary objective is to compare the image generation performance of standard VAR models against those integrated with \ourmethodnospace, focusing on both visual quality, memory utilization and computation efficiency.
Following \cite{var}, we use top-k=600 and CFG=2.0 by default, ensuring a balance between diversity and coherence in generated samples. Results with top-k=900 and CFG=1.5 are explicitly noted when used.
We set the capacity of cache-demanding layers, $C_{\max}$ to exactly equal the size of the last two scales and condensed scales, and the capacity of cache-efficient layers, $C_{\min}$, to the size of the penultimate scale and condensed scales. This configuration enables a systematic analysis of the trade-off between memory consumption and generation quality. Meanwhile, we set $C_{cds}$ as the first two scales.

\paragraph{Evaluation Metrics} The evaluation is conducted on the ImageNet-1K $256\times256$ and $512\times 512$ class-conditioned generation, and object-focused benchmarking framework GenEval\cite{ghosh2023geneval}. 
We adopt standard metrics to measure generation quality: Fréchet Inception Distance (FID) for measuring distributional alignment with real images, Inception Score (IS) for evaluating sample realism and diversity, and Precision and Recall to quantify the trade-off between fidelity and coverage. Memory usage and computation latency is measured on a single NVIDIA A100 (80GB) GPU, and KV cache storage requirements is obtained with PyTorch CUDA profiling tools. \ourmethod is compatible with FlashAttention~\cite{dao2022flashattention}, so we keep using FlashAttention-2 as default with all baselines. 

\paragraph{Results and Analysis} 
Table~\ref{tab:model_performance} presents a comparison of modeling performance and KV cache utilization between the baseline model and \ourmethod across different model sizes. The batch size is fixed at 50 by default. 
We can observe that \ourmethod significantly reduces memory consumption while preserving high generation quality with minimal impact(+0.02 to +0.06 FID) on key evaluation metrics under different configurations. Across different VAR model variants, it achieves a reduction in KV cache utilization ranging from $71.29\%$ to $78.72\%$ (3.48$\times$ to 4.70$\times$ compression). Notably, it enables inference for the largest VAR-d36 model, which would otherwise exceed the memory capacity of a single GPU under standard caching strategies. 




These results confirm that \ourmethod is a highly effective solution for reducing memory overhead while preserving generation quality, making it particularly advantageous for scaling autoregressive vision models within limited hardware constraints. 
\begin{table}[h!]
\centering
\setlength{\tabcolsep}{1.5mm}
\caption{Evaluation of \ourmethod on the Text-to-Image Task Using Infinity-2B on GenEval} 
\label{tab:infinity} 
\begin{tabular}{ccccc}
\toprule
 &\textbf{Two Obj.}$\uparrow$&{\textbf{Position}$\uparrow$} &{\textbf{Color Attr.}$\uparrow$} &{\textbf{Overall}$\uparrow$} \\
\midrule
baseline &\textbf{0.824} &0.277 &0.592 &0.711 \\
+\ourmethod &0.821 &\textbf{0.278} &\textbf{0.608} &\textbf{0.714} \\ 
\bottomrule
\end{tabular}
\end{table}



\begin{table}[ht]
    \centering
    \setlength{\tabcolsep}{3mm}
    \caption{Comparison with other KV cache strategies.}
    \begin{tabular}{lcccc}
        \toprule
        & SWA
        & H2O
        & STA
        & \ourmethod \\
        \midrule
        \textbf{FID$\downarrow$} & 11.61 & 8.80 & 2.81 & \textbf{2.09} \\
        \textbf{IS$\uparrow$}    & 169.94 & 182.8 & 276.92 & \textbf{323.7} \\
        \bottomrule
    \end{tabular}
    \label{tab:amskv_vs_h2o}
\end{table}

\begin{figure}[!h]
    \centering
    \includegraphics[width=0.99\linewidth,trim=100mm 17mm 94mm 25mm,clip]{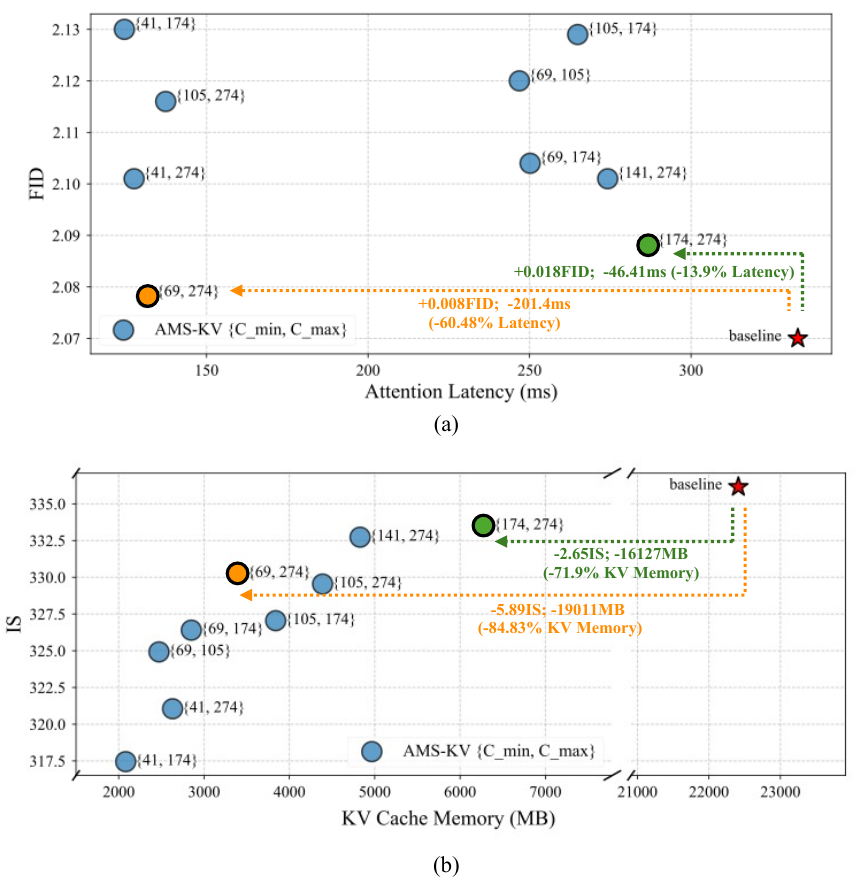}
    \caption{Robustness of \ourmethod across different cache size configurations. (a) FID vs. Attention Latency; (b) IS vs. KV Cache Memory Usage.}
    \label{fig:CminCmaxSweep}
\end{figure}

\paragraph{Robustness of \ourmethod} 
To demonstrate the robustness of \ourmethod beyond intuitive hyperparameters, we explore a broader range of configurations for the tunable parameters $C_{\min}$ and $C_{\max}$. We perform ablation studies by sweeping across valid combinations, as shown in Figure~\ref{fig:CminCmaxSweep}, and observe a variety of memory–performance trade-offs. Notably, most configurations achieve substantial memory savings with minimal generation quality degradation, suggesting significant redundancy in VAR and robustness of \ourmethodnospace. 

\paragraph{Evaluation on Infinity-2B}
We evaluate \ourmethod on Infinity-2B as shown in Table~\ref{tab:infinity}. \ourmethod reduces memory consumption from 28.12GB to 18.02GB($35.6\%\downarrow$) and increases throughput from 0.826 images/s to 0.885 images/s, while slightly improving overall generation quality from 0.711 to 0.714.

\paragraph{Comparison with Other KV Cache work}
We compare the generation performance of a KV cache strategy for LLMs under a fixed compression ratio of 75\%. As shown in Table~\ref{tab:amskv_vs_h2o}, \ourmethod significantly has better generation quality at the same memory budget over SWA~\cite{beltagy2020longformer}, H2O~\cite{h2o}, and STA~\cite{streamingllm}). Meanwhile, \ourmethod operates independently of attention scores, making it fully compatible with memory-efficient attention kernels.

\subsection{Reduced Memory Footprint with \ourmethodnospace}
We further analyze the memory footprint of \ourmethod\space compared to the baseline model as batch size increases—an important factor for the practical deployment of VAR models. Using the same setup as in Section~\ref{sec:main_exp}, we evaluate the VAR-d30 model with batch sizes ranging from 16 to 256. In addition to reporting the KV cache memory footprint, we also measure the change in throughput, defined as the number of images generated per second (\text{img}/s).

\begin{table}[h]
    \centering
    \caption{The Optimization of VAR-d30 Memory Footprints and Throughput Across Batch Sizes.}
    \setlength{\tabcolsep}{2.4mm}{
    \begin{tabular}{cccc}
        \toprule
        \textbf{Batch Size} & \textbf{Model} & \textbf{KV Mem.} & \textbf{Thr.(img/s)} \\
        \hline
        16  & VAR-d30         & 7.17GB  & 18.73 \\
        16  & +\cache            & 2.81GB  & 18.86 \\
        \midrule
        32  & VAR-d30         & 14.34GB & 21.07 \\
        32  & +\cache             & 5.62GB  & 21.10 \\
        \midrule
        64  & VAR-d30         & 28.69GB & 22.21 \\
        64  & +\cache             & 11.23GB & 22.32 \\
        \midrule
        128 & VAR-d30         & \texttt{OOM}   & \texttt{OOM}     \\
        128 & +\cache             & 22.47GB & 23.00 \\
        \midrule
        256 & VAR-d30         & \texttt{OOM}   & \texttt{OOM}     \\
        256 & +\cache             & 44.94GB & 23.88 \\
        \bottomrule
    \end{tabular}
    }
    \label{tab:throughput}
\end{table}

\paragraph{Results and Analysis:} 
Table~\ref{tab:throughput} shows that as batch size increases, the KV cache size grows rapidly and eventually dominates overall memory utilization. 
Notably, at batch sizes of 128 and 256, the baseline model of VAR-d30 encounters Out-Of-Memory(\texttt{OOM}) failure, making inference infeasible. In contrast, \ourmethod significantly reduces KV cache requirements, allowing inference under 4$\times$ batch size.

Despite the substantial memory savings, \ourmethod yields only a modest throughput improvement of 7.5\%. This is expected, as the attention computation latency is not dominant factors compared to FFNs in VAR models, and thus does not significantly constrain overall throughput as memory usage does. 
However, the ability of \ourmethod to support inference at high batch sizes without memory overflows highlights its practical advantage for real-world deployment, where large-batch processing is often essential for efficiency.






\section{Conclusion}
We address the challenge of KV caching inefficiency in visual autoregressive models, a major bottleneck for scalability. While VAR models achieve state-of-the-art generative performance with low latency, their growing KV cache footprint limits practical deployment. To mitigate this, we propose \ourmethodnospace, an adaptive multi-scale KV caching strategy that significantly reduces memory consumption while preserving model performance.
Experiments show that \ourmethod reduces KV cache utilization by up to $78.72\%$, enabling inference at 4 $\times$ batch sizes and preventing OOM failures. Despite this reduction, it maintains highly competitive image generation quality. Our analysis further reveals that early layers require larger caches, while later layers tolerate compression, offering valuable insights into efficient cache management. By alleviating KV cache overhead, \ourmethod enhances the scalability of VAR models, facilitating deployment in memory-constrained settings. We hope this work inspires future research on optimizing AR-based vision models for improved efficiency and scalability.


\section{Acknowledgement}
This material is based upon work supported by the National Science Foundation under Grants No. 2310170.
\bibliography{aaai2026}
\appendix
\section{Appendix}
\section{A\quad Additional Experiment Results}

\subsection{The Ablated Effects of Condensed Scale $C_{\text{cds}}$}
A critical architectural choice in \ourmethod is selecting the number of initial scales to retain as condensed scales in the KV cache. These condensed scales remain cached throughout the entire inference process, while subsequent intermediate scales are dynamically flushed out and appended by more local scales. 
The main paper discusses the condensing phenomenon in which the first and second scales capture critical structural information and exhibit the highest attention density and stabilizes multi-scale generation by preserving structural priors with a few number of tokens. 
To evaluate the impact of the number of condensed scales ($C_{\text{cds}}$), we conduct an empirical study using the VAR-d30 model. 

\begin{table}[h]
    \centering
    \small
    \caption{Effect of the Size of Condensed Scales ($C_{\text{cds}}$) on Generation Quality}
    \setlength{\tabcolsep}{2.7mm}{
    \begin{tabular}{l | c c | c c}
        \toprule
        \textbf{CDS} & \textbf{FID$\downarrow$} & \textbf{IS$\uparrow$} & \textbf{Precision$\uparrow$} & \textbf{Recall$\uparrow$} \\
        \midrule
        CDS = $\emptyset$  & 2.48 & 298.35 & 0.80 & 0.59 \\
        CDS = $\{0\}$  & 2.16 & 333.04 & 0.82 & 0.57 \\
        CDS = $\{0,1\}$  & 2.09 & 333.51 & 0.82 & 0.58 \\
        Baseline & 2.07 & 336.15 & 0.82 & 0.58 \\
        \bottomrule
    \end{tabular}
    }
    \label{tab:cds_effect}
\end{table}

Table~\ref{tab:cds_effect} quantitatively demonstrates the effect of including condensed scales. Removing all condensed scales results in a substantial performance drop: FID increases from 2.07 to 2.48, and IS drops from 336.15 to 298.35. In contrast, caching the first scale ($r_1$) improves FID to 2.16 and IS to 333.04, while caching both $r_1$ and $r_2$ with only 5 tokens further improves FID to 2.09 and IS to 333.51. These results confirm that including additional initial scales as condensed scales effectively mitigates quality degradation.

\begin{figure}
    \centering
    \includegraphics[width=0.99\linewidth]{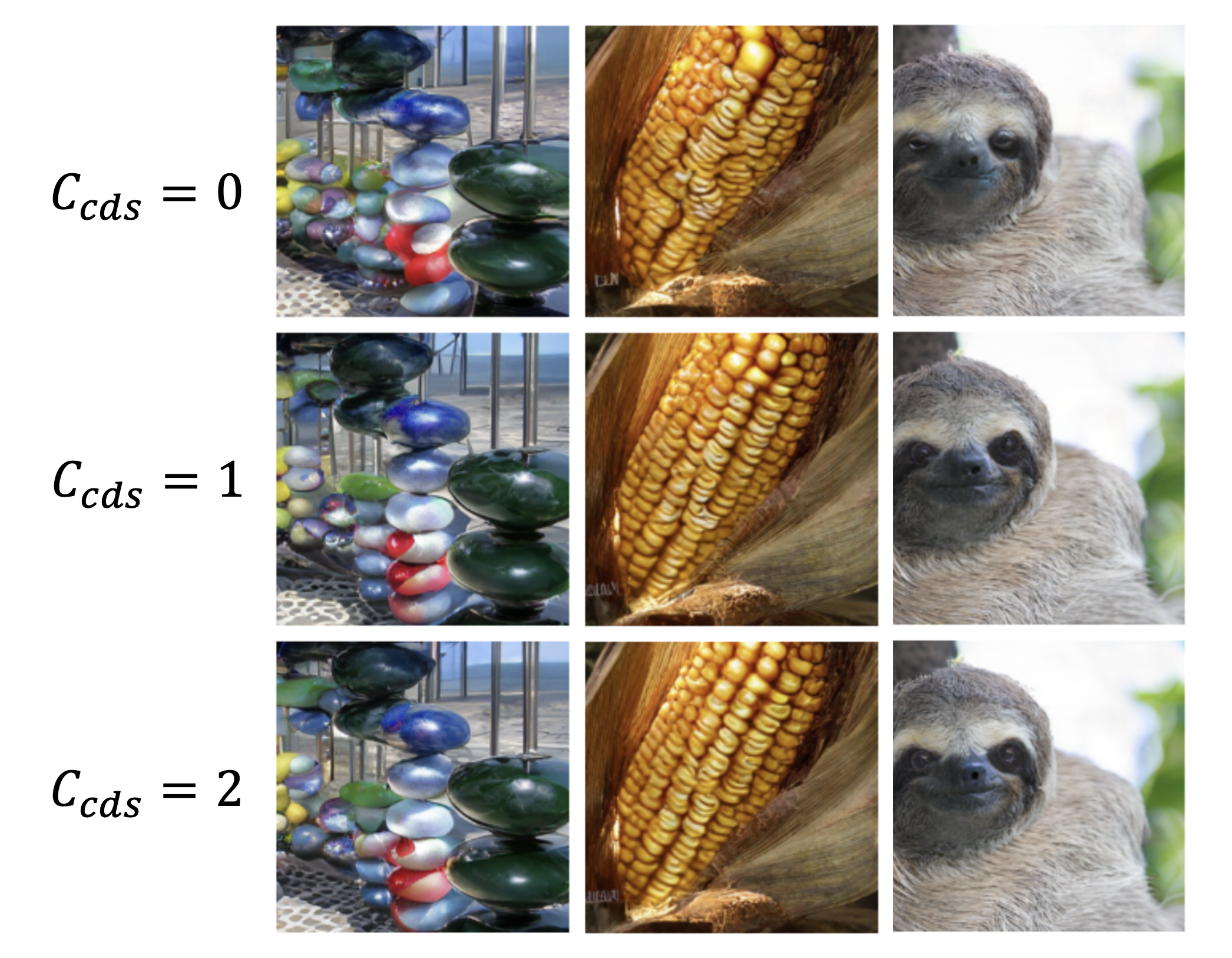}
    \caption{Illustrative examples comparing the effect of different numbers of condensed scales on image generation quality.}
    \label{fig:CDS_effects}
\end{figure}

Qualitative results in Figure~\ref{fig:CDS_effects} further support this conclusion. Without condensed scales, generated images often distort global structure and overemphasize residual details. In contrast, progressively incorporating more condensed scales significantly enhances structural coherence and visual sharpness, underscoring the importance of caching condense scales in preserving generation quality.

\subsection{Impact of \ourmethod on Domain Shift}
\begin{table}[h!]
\centering
\caption{Evaluation of FID$\downarrow$ on Diff. Domains on ImageNet}
\label{tab:domain_shift}
\begin{tabular}{ccc} 
\toprule
 & \textbf{Landscape-type} & \textbf{Portraits-type} \\
\midrule
baseline & 57.89 & 62.60 \\
AMS-KV & 57.49(-0.4$\downarrow$) & 62.61(+0.01$\uparrow$) \\
\bottomrule
\end{tabular}
\end{table}
We evaluate the quality of generated images using the VAR-d30 model on both portrait and landscape classes from the ImageNet dataset, as summarized in Table~\ref{tab:domain_shift}. 
\ourmethod demonstrates robust performance across both domains. Notably, landscape-type images exhibit lower sensitivity to the application of \ourmethodnospace. Specifically, \ourmethod reduces the FID for landscape classes from 57.89 to 57.49, while for portrait classes, the FID remains nearly unchanged (62.60 to 62.61), indicating stable performance under domain variation.

\section{B\quad Qualitative Results for \ourmethodnospace-enhanced VAR models}
\begin{figure*}[!t]
    \centering
    \includegraphics[width=0.99\linewidth, trim=30mm 140mm 70mm 90mm, clip]{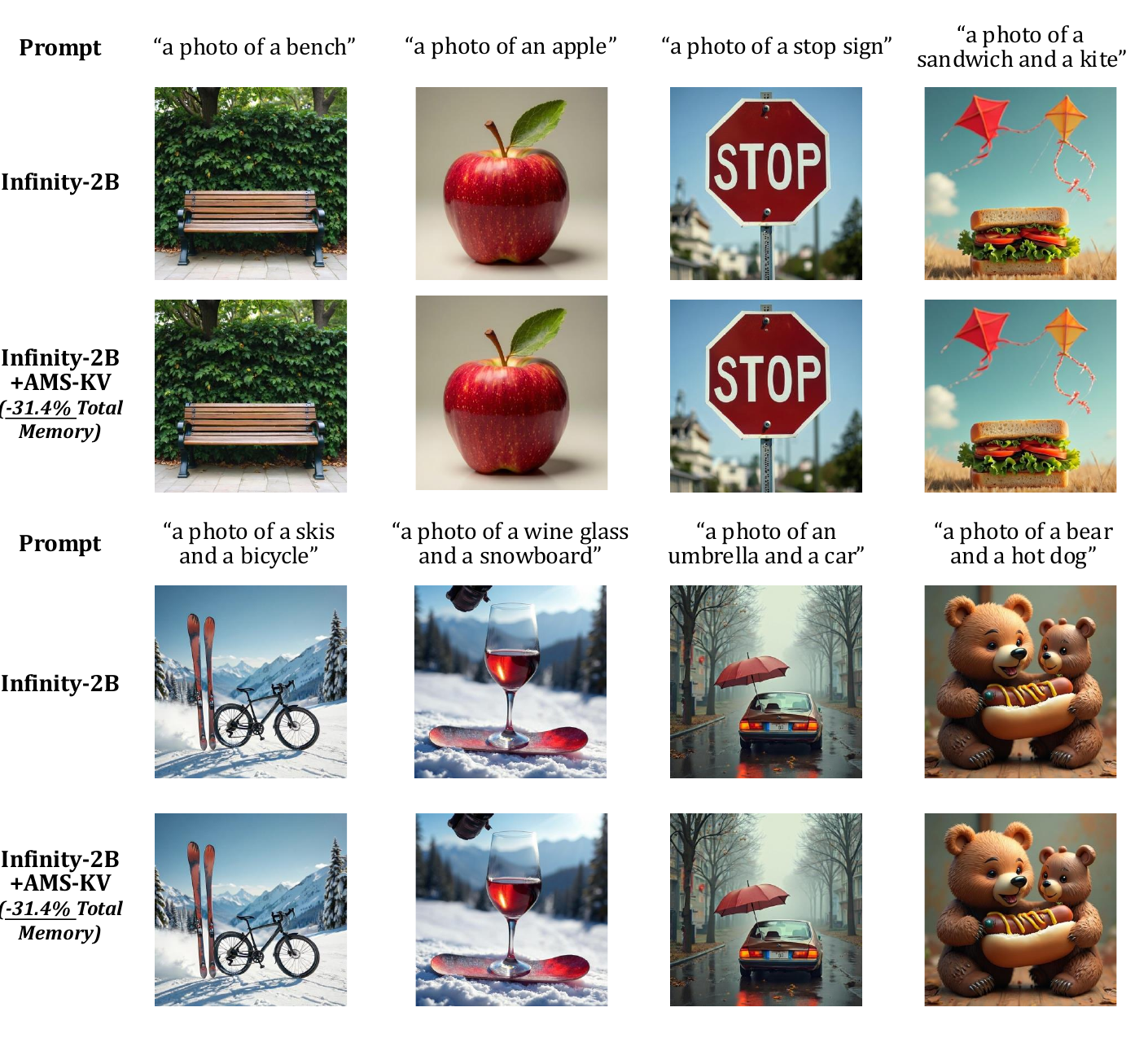}
    \caption{Qualitative Comparison between Synthesized Images of Infinity-2B baseline and with \ourmethod added.}
    \label{fig:Infinity_examples}
\end{figure*}

This section presents additional qualitative results from image generation using the Infinity-2B model. We use prompts from GenEval~\cite{ghosh2023geneval}, and we set the classifier-free guidance (CFG) scale to 4.0. As shown in Figure~\ref{fig:Infinity_examples}, \ourmethod achieves a 31.4\% reduction in total memory usage, enabling inference on low HBM-capacity GPUs without generation quality degradation.

\section{C\quad The Capacity of Detail Preserving}
We further assess low-level metrics including PSNR, SSIM, and LPIPS, treating frames generated with full-KV caching as the ground truth. As reported in Table~\ref{tab:amskv_lowlevel}, \ourmethod consistently attains higher PSNR and SSIM together with lower LPIPS than alternative KV caching schemes, even in a fully tuning-free setting. These results indicate that the synthesized images produced by \ourmethod more closely match the full-KV outputs in terms of pixel-wise accuracy, structural similarity, and perceptual quality. Consequently, the proposed cache compression preserves more low-level visual details and higher human-perceived consistency with full-KV decoding, while substantially reducing the KV memory footprint.

\section{D\quad Formalization of the memory saving}
We define the KV cache memory saving (in MB) in Equation~\ref{eq:kv_mem_saving}, where $B$ denotes the batch size, $H$ is the number of attention heads, and $D$ is the head dimension. $T_s$ represents the number of tokens at scale $s$, $N$ denotes the total number of layers, and $S$ represents the set of all scales. $S_{\mathrm{CD}}$ and $S_{\mathrm{CE}}$ denote the sets of scales retained in the KV cache at cache-demanding and cache-efficient layers, respectively, while $N_{\mathrm{CD}}$ and $N_{\mathrm{CE}}$ represent the number of cache-demanding and cache-efficient layers, respectively.
\begin{table}[h]
    \centering
    \setlength{\tabcolsep}{3.5mm}
    \caption{Comparison with SWA~\cite{beltagy2020longformer} and STA~\cite{streamingllm} on detail preservation, based on low-level metrics computed with full-KV caching outputs as ground truth.}
    \begin{tabular}{lccc}
        \toprule
        \textbf{Method} & \textbf{PSNR$\uparrow$} & \textbf{SSIM$\uparrow$} & \textbf{LPIPS$\downarrow$} \\
        \midrule
        SWA        & 16.61 & 0.403 & 0.287 \\
        STA        & 18.50 & 0.496 & 0.198 \\
        \ourmethod & \textbf{21.92} & \textbf{0.669} & \textbf{0.112} \\
        \bottomrule
    \end{tabular}
    \label{tab:amskv_lowlevel}
\end{table}
\begin{equation}
\label{eq:kv_mem_saving}
\begin{aligned}
&\text{KV Mem Saving (MB)} = \frac{16BHD}{1024^{2}} \times \\
&\Biggl[\Biggl(\sum_{s \in S} T_s \Biggr) N - \Biggl(\sum_{s \in S_{\mathrm{CD}}} T_s \Biggr) N_{\mathrm{CD}} - \Biggl( \sum_{s \in S_{\mathrm{CE}}} T_s \Biggr) N_{\mathrm{CE}} \Biggr].
\end{aligned}
\end{equation}


\section{E\quad Robustness of Similarity Thresholds and Other Hyperparameters}
The similarity threshold $\theta$ in AMS-KV is not arbitrarily chosen. Intuitively, setting $\theta$ too large would classify almost all layers as cache-demanding, while setting it too small would classify almost all layers as cache-efficient. To avoid such degenerate regimes, we perform a one-time offline profiling step to determine a meaningful operating range of $\theta$. On VAR-d30, we observe that the performance is stable within this range. Specifically, we obtain FID/IS scores of 2.21/294.15 for $\theta=-0.01$, 2.18/295.03 for $\theta=-0.009$, 2.15/295.71 for $\theta=-0.008$, and 2.16/295.56 for $\theta=-0.007$. These results indicate that \ourmethod is robust to moderate variations of the similarity threshold, without requiring any per-model or per-dataset tuning.


\section{F\quad Condensed and Least Recently Used Cache}
In principle, CLRU is a cache eviction policy that behaves like a FIFO scheme while always keeping the condensed scales resident. As shown in Algorithm~\ref{alg:clru}, when the size of $\mathcal{C}_{i}$ is within the budget $C{bgt}$, CLRU leaves the cache unchanged. If the size exceeds $C_{bgt}$, it retains tokens from $C_{\text{cds}}$ condensed scales and keeps the remaining $C_{\text{bgt}} - C_{\text{cds}}$ most recent local tokens, evicting the others.

\begin{algorithm}[t!]
\caption{Cache with Eviction Policy of Condensed and Least Recently Used (CLRU)}
\label{alg:clru}
\begin{algorithmic}[1]
\Function{\texttt{CLRU}}{$\mathcal{C}_i, C_{bgt}, C_{\text{cds}}$}
    \If{$|\mathcal{C}_i| \le C_{\text{bgt}}$}
        \State \Return $\mathcal{C}_i$ \Comment{no rolling if within budget}
    \EndIf
    \State $\mathcal{C}^{\text{cds}} \gets \mathcal{C}_i[:C_{\text{cds}}]$  \Comment{keep condensed scales}
    \State $k \gets C_{bgt} - C_{\text{cds}}$
    \State $\mathcal{C}^{\text{tail}} \gets \mathcal{C}_i[-k:]$ \Comment{keep non-evicted local scales}
    \State \Return $\mathcal{C}^{\text{cds}} \cup \mathcal{C}^{\text{tail}}$
\EndFunction
\end{algorithmic}
\end{algorithm}

\section{G\quad Potential Societal Impacts}
This work identifies unique properties of KV caching in visual autoregressive (VAR) model families, and contributes to improving the scalability and deployment feasibility of VAR models by addressing memory inefficiencies in KV caching. 
By significantly reducing KV memory consumption without compromising generation quality, \ourmethod enables memory-efficient inference on modern GPUs. Combined with the inherently low-latency nature of VAR models\cite{var, infinity} compared to diffusion models, this advancement facilitates broader deployment of generative models in resource-constrained environments, including edge computing platforms, low-power devices, and regions where high-memory hardware is inaccessible or cost-prohibitive.


\end{document}